%% file: 0_main_arxiv.tex
\documentclass[a4paper, 10pt, conference]{ieeeconf}      % Use this line for a4
\usepackage{FG2024}
 \usepackage{colortbl}
 \usepackage{multirow}
 \usepackage{float}
\usepackage{subcaption}
\usepackage{graphicx}
\usepackage{footnote}
\usepackage{amsmath}
\usepackage{url}
\usepackage[colorlinks=true]{hyperref}
\usepackage{times}  % Ensures Times New Roman is used and embedded
\usepackage[T1]{fontenc}

\FGfinalcopy % *** Uncomment this line for the final submission

\IEEEoverridecommandlockouts                              % This command is only
\title{\Large \bf
Benchmarking Skeleton-based Motion Encoder Models for Clinical Applications: \\ Estimating Parkinson's Disease Severity in Walking Sequences
}

\author{\parbox{16cm}{\centering
    {\large Vida Adeli$^{1,2}$, Soroush Mehraban$^{1,2}$, Irene Ballester$^{3}$, Yasamin Zarghami$^{1,2}$, Andrea Sabo$^{2}$, Andrea Iaboni$^{1,2,3}$ and Babak Taati$^{1,2}$}\\
    {\normalsize
    $^1$ University of Toronto, CA,
    $^2$ KITE - Toronto Rehabilitation Institute – University Health Network, CA\\
    $^3$ Vienna University of Technology, AUST,
    $^4$ Center for Mental Health – University Health Network, CA\\
    }}
    \thanks{This work was supported by the Walter and Maria Schroeder Institute
for Brain Innovation and Recovery; KITE | Toronto Rehabilitation
Institute -- University Health Network; the Canadian Institute of Health Research (CIHR), the Natural Sciences and Engineering Research Council; 
Alzheimer's Association  \&
Brain Canada; Data Sciences Institute, University of Toronto;
and
AMS Healthcare Fellowship in Compassion and Artificial
Intelligence.}% <-this % stops a space
}

\usepackage{fancyhdr}
\begin{document}

\ifFGfinal
\thispagestyle{empty}
\pagestyle{empty}
\else
\author{Anonymous FG2024 submission\\ Paper ID \FGPaperID \\}
\pagestyle{plain}
\fi
\maketitle

%%%%%%%%%%%%%%%%%%%%%%%%%%%%%%%%%%%%%%%%%%%%%%%%%%%%%%%%%%%%%%%%%%%%%%%%%%%%%%%%
\thispagestyle{fancy}
\begin{abstract}
Parkinson's Disease is a degenerative disorder for which precise motor assessment is critical.
This study investigates the application of general human motion encoders trained on large-scale human motion datasets for analyzing gait patterns in PD patients. 
Although these models have learned a wealth of human biomechanical knowledge, their effectiveness in analyzing pathological movements, such as parkinsonian gait, has yet to be fully validated.
We propose a comparative framework and evaluate six pre-trained state-of-the-art human motion encoder models on their ability to predict the Movement Disorder Society -- Unified Parkinson's Disease Rating Scale (MDS-UPDRS-III) gait scores from motion capture data. We compare these against a traditional gait feature-based predictive model in a recently released large public PD dataset, including PD patients on and off medication. The feature-based model currently shows higher weighted average accuracy, precision, recall, and F1-score. Motion encoder models with closely comparable results demonstrate promise for scalability and efficiency in clinical settings. This potential is underscored by the enhanced performance of the encoder model upon fine-tuning on PD training set. Four of the six human motion models examined provided prediction scores that were significantly different between on- and off-medication states. This finding reveals the sensitivity of motion encoder models to nuanced clinical changes, emphasizing their potential utility in clinical settings. It also underscores the necessity for continued customization of these models to better capture disease-specific features, thereby reducing the reliance on labor-intensive feature engineering. Lastly, we establish a benchmark for the analysis of skeleton-based motion encoder models in clinical settings. To the best of our knowledge, this is the first study to provide a benchmark that enables state-of-the-art models to be tested and compete in a clinical context. Codes and benchmark leaderboard are available at 
\href{https://github.com/TaatiTeam/MotionEncoders_parkinsonism_benchmark.git}{code}.
% \small{\url{http://blinded/}}.

%The results showed that the human motion priors learned from general datasets require further refinement and integration of clinical knowledge to effectively interpret the gait abnormalities present in PD.

% The feature-based approach extracts traditional gait features, while the motion prior-based pathway leverages pre-trained models on general motion to directly estimate UPDRS-III scores. Our comparative analysis reveals that while traditional gait features provide a solid baseline, the motion prior-based model, through fine-tuning, shows promise for improved accuracy.

\end{abstract}

%%%%%%%%%%%%%%%%%%%%%%%%%%%%%%%%%%%%%%%%%%%%%%%%%%%%%%%%%%%%%%%%%%%%%%%%%%%%%%%%
\input{1_Introduction}
\input{2_RelatedWork}
\input{3_Method}
\input{4_Experiments}

%%%%%%%%%%%%%%%%%%%%%%%%%%%%%%%%%%%%%%%%%%%%%%%%%%%%%%%%%%%%%%%%%%%%%%%%%%%%%%%%
\section{LIMITATIONS AND FUTURE WORK}
While this study makes significant strides in applying motion encoder models to parkinsonian gait analysis, it is not without limitations, which also point toward future research directions.

One key limitation lies in the dataset's scope, particularly concerning the ground-truth labels being assigned per participant's medication state (ON or OFF), not for each individual walk. This contrasts with our models that predict gait characteristics for each walk. As PD symptoms are known to fluctuate, the dataset’s broad labeling approach may not accurately capture the exact parkinsonian score in each walk.
Further, the majority of the data represents PD patients with mild to moderate symptoms (UPDRS scores 0 to 2), limiting our ability to generalize findings to more severe cases. Future studies should aim to include a broader spectrum of PD severity, particularly focusing on higher UPDRS scores, to validate and potentially enhance the model's efficacy across all stages of the disease. We note that a UPDRS-gait score of 3 is defined as requiring an assistance device (e.g. walking stick or walker) for safe walking, and a score of 4 is defined as inability to walk at all or only with another person’s assistance. Therefore, expanding datasets to include walking sequences with these severe scores is particularly challenging.

Another limitation is the reliance on pre-trained models, which, despite fine-tuning, may not fully capture the intricate details of parkinsonian gait patterns. Currently, the absence of large-scale clinical datasets for PD restricts the effective training or sufficient fine-tuning of such large models. Future research could explore the development of models trained explicitly on PD gait data, potentially offering more accurate and clinically relevant insights.
To address this gap, future efforts could focus on the creation or augmentation of PD-specific datasets. Considering the challenges in collecting extensive clinical data, one viable approach is the generation of synthetic data, infused with clinical knowledge pertinent to PD.

% Additionally, the models' varying ability to distinguish between ON and OFF medication states in PD patients highlights the need for personalized model selection in clinical applications. Future research should delve deeper into personalized AI solutions in healthcare, tailoring models to individual patient profiles for more precise treatment and monitoring.

The study also reveals that while motion encoder models like PoseFormerV2-Finetuned demonstrate promise, they are not superior to traditional feature-based methods. This finding suggests that future work should not only continue to refine these advanced models, but also explore integrative approaches to take advantage of expert knowledge and common clinical practices to improve performance.

The performance of the PoseFormerV2-Finetuned model, while generally effective, reveals certain limitations, particularly in distinguishing between mild and severe gait impairments. This aspect is critical for clinical use and suggests areas for further refinement.

Lastly, the study’s focus on gait analysis leaves other PD symptoms, such as tremor or rigidity, unexplored. Subsequent studies could expand the application of motion encoder models to these other symptoms, offering a more comprehensive toolset for PD assessment and monitoring. %Developing personalized models based on individual patient profiles, i.e. focusing on the  history and gait pattern variations of each patient, could be another interesting direction for future research.

% The current study’s findings are derived from a specific dataset, which may limit the generalizability of the results to all Parkinson’s patients or other diseases. Future studies should aim to validate these findings across a broader demographic and range of conditions.

\section{CONCLUSIONS}

This research aimed to assess the effectiveness of advanced motion encoder models in analyzing parkinsonian gait. 
Our findings highlight that while these models require specific adjustments for clinical use, they hold significant promise. %, particularly in distinguishing gait changes related to PD medication states.
The study's results underscore the crucial role of model selection and fine-tuning, particularly in medical contexts like PD gait analysis. While the PoseFormerV2-Finetuned model, specifically adapted for this task, showed high performance, it, along with other motion encoder models, did not surpass the feature-based approach. This highlights substantial opportunities for improvement, particularly in domain adaptation. However, the inherent adaptability of motion encoders, not being task-specific, positions them favorably for various downstream tasks, unlike traditional feature-based methods that require extensive task-specific feature engineering.
Our statistical analysis reveals the efficacy of motion encoder models, particularly when fine-tuned, in detecting medication-related gait changes in PD patients. This ability to discern clinically significant gait variations underscores the importance of such models in developing tools for therapy monitoring and adjustment in PD. %However, the varied performance across models also indicates the necessity for tailored model selection in clinical use, based on the specific requirements for sensitivity in differentiating ON and OFF medication states.

In conclusion, our study demonstrates the potential of leveraging pre-existing knowledge in motion encoder models, enhanced with additional data, for developing specialized models for PD score estimation. This approach not only enhances the accuracy and relevance of PD assessments but also opens up new avenues for AI application in clinical settings, emphasizing the need for nuanced, patient-specific solutions.

{\small
\bibliographystyle{ieee}
\bibliography{egbib}
}

\end{document}

%% file: 1_Introduction.tex
\vspace{10pt}
\section{INTRODUCTION}
\vspace{5pt}

\pagestyle{fancy}
% Gait analysis is essential to understand biomechanical complexities and diagnose degenerative diseases, particularly Parkinson’s Disease (PD). By analyzing movement, clinicians gain valuable insights into neurodegenerative disorders, enabling early detection and customized intervention strategies. PD, notably, has a profound effect on motor functions, with gait disturbances being a primary manifestation \cite{matsushita2021recent}.

Parkinson's Disease (PD) is a progressive neurological disorder that affects motor function, often leading to severe impairment in movement and gait~\cite{matsushita2021recent}. Accurate assessment of gait disturbance in people with PD is crucial for diagnosis, monitoring of the disease progression, and response to medications. By observing movement impairments and their changes over time, clinicians can adjust medication dosage or prescribe customized intervention strategies.

Traditionally, clinicians rely on observational techniques and patient-reported measures, such as the Movement Disorder Society -- Unified Parkinson’s Disease Rating Scale (MDS-UPDRS)~\cite{goetz2008movement}, which rates symptom severity in the movements of different body parts on a 0-4 scale. However, these methods are subjective and may not capture the full extent of motor impairments, including
% . While objective markers 
% like bradykinesia and akinesia 
% exist, subjective symptoms such as 
tremor, rigidity, and postural instability~\cite{parkinsonMarkers}. Gait disturbances in PD are particularly challenging to monitor due to their fluctuations in response to treatment, often eluding conventional in-clinic assessments. As a result, there is a growing need for objective and validated methods to assess PD symptom severity, particularly in the context of gait analysis.

Advances in human motion encoder models in computer vision and biomechanics offer an opportunity to address this requirement. These models, originally developed for general human motion data, have the potential to transform raw gait data into detailed, quantifiable metrics. Yet, their application in a clinical setting, including PD assessment, remains largely unexplored. This gap in research signifies not just an opportunity, but a necessity to explore the utility of these advanced technologies in a clinical context.

Traditionally, gait analysis in PD has been confined to manual observation or rudimentary computational methods focusing on a limited set of gait features \cite{sabo2020assessment, rupprechter2021clinically, chavez2022vision}. While these methods have provided valuable insights, they fall short in capturing the complex and subtle motor deficits characteristic of PD. In contrast, recent advancements in human motion encoder models, trained and validated on large datasets comprising healthy individuals, present an opportunity to revolutionize this domain. These models, having demonstrated high performance in general motion capture and analysis, have not yet been systematically evaluated or adapted for clinical settings, particularly for conditions characterized by atypical movement patterns like PD.

Our study focuses on this intersection, aiming to harness the power of cutting-edge motion encoder models for the clinical evaluation of PD. We delve into the capabilities of these models to interpret complex gait patterns in PD patients, examining whether they can effectively translate skeletal movement data into clinically meaningful insights. We explore whether these advanced technologies can go beyond the capabilities of traditional feature-based methods in interpreting skeletal movement data. %, thereby offering a more objective and comprehensive assessment of PD severity. 
This study aims to bridge this gap by comprehensively assessing six recent human motion encoder models in their ability to encode and differentiate gait patterns specific to PD. The focus is not merely on the general accuracy of these models, but on their sensitivity to disease-specific gait abnormalities, often overlooked or misinterpreted as noise in standard motion analysis. By doing so, this study seeks to establish a benchmark for the application of these sophisticated models in a clinical context, offering a pathway for enhanced patient assessment and, ultimately, better management of Parkinson's disease. We also perform an analysis of patients' gait patterns in both ON and OFF medication states, using statistical analysis to underscore the models' ability to discern clinically significant changes due to medication.

% \subsection{Automated Estimation of Parkinsonism's Severity Score}

% This research utilizes a public motion capture dataset \cite{pdDataset}, encompassing gait dynamics of individuals with PD, to bridge the gap between raw data and clinical insights.

% We introduce a new approach by applying state-of-the-art human motion encoder models to gait analysis and PD severity estimation in clinical settings. This initiative aims to evaluate the effectiveness and applicability of these advanced models in real-world clinical scenarios. We contrast this with traditional gait feature extraction techniques.

% Our approach involves predicting Parkinsonism scores, specifically the UPDRS-III-gait score, to enhance our comprehension of human motion affected by neurodegenerative disorders. By doing so, we aim to facilitate timely interventions and personalized care.

We refer to the MDS-UPDRS Part III, gait subscore for classifying PD gait impairments, hereafter referred to as UPDRS-gait. The UPDRS-gait scores range from 0, indicating no detectable gait impairment, to 4, where patients are unable to walk independently. Scores 0 to 2, reflect a gradual decline in mobility, often manifested through diminished stride length and foot lift, making these gait patterns subtly different from each other and challenging to distinguish. On the other hand, gait patterns with scores above 2 are more discernible as they are characterized by a significant loss of independent walking ability \cite{goetz2019mds}. Consequently, PD public datasets predominantly feature scores ranging from 0 to 2, reflecting these more subtle and common variations in gait, which present a significant challenge for accurate classification and analysis.
In summary, the main contributions of our paper are:

\begin{itemize}
    \item We conduct a thorough evaluation of the effectiveness of six general motion encoder models~\cite{sabo2022estimating, zhang2022mixste, motionagformer2024, motionbert2022, Zhao_2023_CVPR, POTR} in clinical settings, specifically for PD severity estimation. Our findings reveal that, in their pre-trained form, most of these models may not be suited for clinical use, highlighting the need for specialized adaptations. One model (PoseFormerV2~\cite{Zhao_2023_CVPR}), however, displayed comparable results to the feature-based approach after fine-tuning.    
    \item Results from our analysis based on the ON and OFF medication states demonstrate that motion encoder models, even without fine-tuning, already exhibit a promising capacity %to outperform traditional feature-based methods
    in terms of sensitivity to clinically significant changes, suggesting their potential in clinical applications.
    % However, statistical analysis further shows that the motion encoder model can capture more relevant information, such as the significant differences in the patient's PD scores when they are on or off Parkinson's medication, compared to conventional feature-based methods.
    % Further, our experiments indicate that these models show great potential once they are appropriately adapted and fine-tuned for the specific challenges of PD assessment.

    % We also assess the potential for customizing these models to suit clinical needs. Our research demonstrates that with the integration of additional data and domain knowledge, these models can be effectively tailored to outperform traditional feature-based methods.
    % \item We compare these motion encoder models against a traditional gait feature-based method. Through this comparative study, we determine which motion encoder models provide a more accurate and clinically relevant analysis of PD gait patterns than feature-engineered methods.
    \item We are the first to use the public PD dataset~\cite{pdDataset} for estimating PD severity scores. After adapting the best performing motion encoder to parkinsonian movement patterns via fine-tuning, we achieve significant results with a weighted average F1-score of 62\% in a Leave-One-Subject-Out-Cross-Validation (LOSOCV) framework and statistically significant differences between ON and OFF medication states (p-value of 0.0052). Additionally, the feature-based approach yielded an even higher weighted average F1-score of 66\%.
    % showcasing the potential for human motion models to capture clinically relevant information. %more effectively than feature-based methods, as confirmed by the statistical analysis we conducted based on medication status of patients.
    \item Lastly, we establish a benchmark for the application of skeleton-based motion encoder models in clinical settings, providing the human motion analysis research community with the opportunity to not only achieve breakthroughs in general-purpose models but also to compete and evaluate their models in a clinical context. This fosters a bridge between research and practical clinical applications, crucial for guiding future advancements in medical diagnostics and therapeutic interventions for movement disorders.
\end{itemize}

%% file: 2_RelatedWork.tex
\section{RELATED WORK}
% \vspace{5pt}

The study of gait patterns in PD has evolved with the integration of interdisciplinary research encompassing neurology, biomechanics and computer science. Early investigations relied on clinical observations, later standardized by the UPDRS scale to quantify motor symptoms~\cite{noyce2016prediagnostic, UPDRS}.

Following rapid advances in machine learning, research in this area for gait analysis~\cite{liu2022symmetry, teepe2022towards} and its use for PD severity prediction gained momentum. Despite substantial advances, human motion encoders, which have shown potential in several downstream tasks, remain largely unexplored in clinical contexts, including their application in PD assessment.

% To contextualize our research, we begin by reviewing previous studies on predictive models for the Parkinsonian score, followed by an introduction to human motion encoders.

\subsection{PD Severity Estimation}

%Intro and different sensors
In previous research, machine learning algorithms for classifying PD gait severity into different stages have effectively processed gait data collected from different sensors. These sensors include wearables~\cite{henderson2016rivastigmine, cuzzolin2017metric}, camera-based motion capture (Mocap) systems~\cite{park2021classification}, force sensors~\cite{el2020deep, balaji2021automatic, naimi20231d}, Inertial Measurement Units (IMU)~\cite{caramia2018imu, mirelman2021detecting}, and various vision-based approaches. Vision-based approaches employ RGB cameras and depth sensors, either used individually~\cite{guayacan2021visualising, cao2021video, verlekar2018automatic, liu2022quantitative, guo20193} or in combination, such as in Microsoft's Kinect~\cite{eltoukhy2017microsoft, geerse2018assessing, dranca2018using, khokhlova2019normal}.

%Focus: skeleton and joint trajectory

Regardless of the sensor type—Mocap systems, Microsoft Kinect~\cite{khokhlova2019normal}, or cameras with posterior pose estimation (e.g.,~\cite{sabo2020assessment, lu2021quantifying, lu2020vision})—numerous approaches rely on joint trajectories (skeleton sequences) as their input. This preference stems from the established effectiveness of joint trajectories in capturing subtle nuances associated with pathological PD gait~\cite{sabo2022estimating, tian2024cross}.

%Hand-crafted gait features + classifiers

Models using skeleton data can be categorized into two groups: 1) those that extract hand-crafted gait features based on prior knowledge and 2) end-to-end deep learning architectures designed for direct score prediction. In the first category, hand-crafted features are used as inputs to train machine learning algorithms such as support vector machines (SVMs), decision trees, and deep classification algorithms to predict the score. In this line,  Khokhlova et~al.~\cite{khokhlova2019normal} proposed a bidirectional Long Short-Term Memory (LSTM) network to evaluate gait symmetry by extracting features from 3D lower limb flexion dynamics. Similarly, Li et~al.~\cite{Li2018,li2018automated} used Convolutional Pose Machines~\cite{wei2016convolutional} to extract movement trajectory features from videos for training random forests to detect and estimate the severity of parkinsonism and levodopa-induced dyskinesia (LID). In~\cite{sabo2020assessment}, sixteen 3D and eight 2D gait features were extracted from natural gait videos and analyzed using a multivariate ordinal logistic regression method to detect PD. While not yet validated for its use in PD severity estimation, noteworthy recent works for gait feature extraction include TransformingGait~\cite{TransformingGait} and Pose2Gait~\cite{Pose2gait}, employing transformers and convolutions to extract hand-crafted features by mapping joint trajectories and pose sequences, demonstrating proficiency in capturing intricate temporal dependencies in gait patterns.

%Why not handcrafted methods
%However, handcrafted methods have been proven to have limited ability to extract generalsable and distinctive representations~\cite{nanni2017handcrafted,el2020deep,}. 
\begin{figure*}[t]
  \centering
  \includegraphics[width=0.9\linewidth]{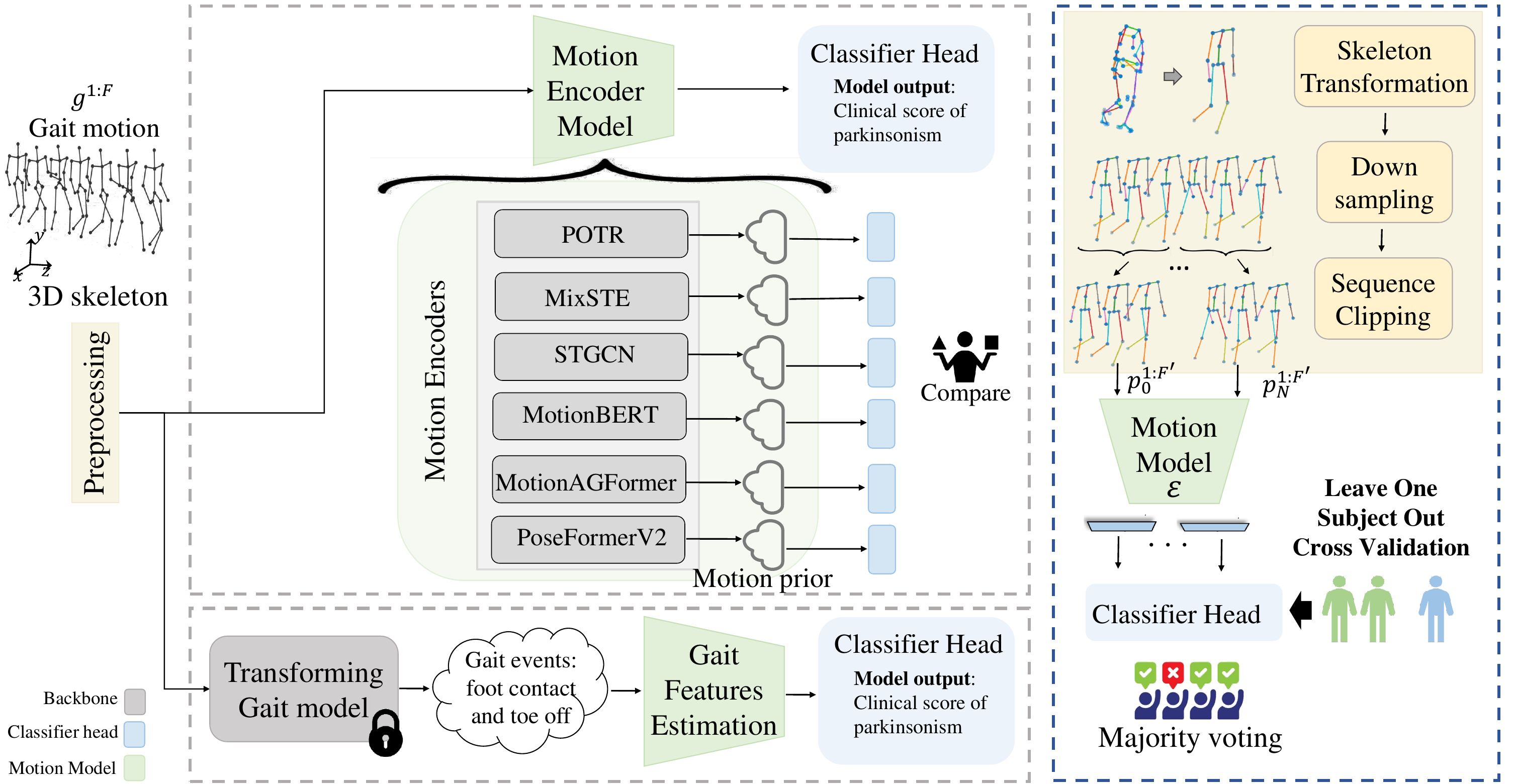}
  \caption{Overview of the framework for UPDRS-gait score estimation. (top) employs motion encoder models to predict UPDRS-gait scores directly from gait motion, while (bottom) uses a feature-based approach for score estimation from gait features estimated from motion data. (right) encapsulates the full pipeline from data preprocessing to model evaluation, employing LOSOCV and majority voting to validate and compare the effectiveness of both approaches against clinical UPDRS-gait scores.}
  \label{fig:diagram}
\end{figure*}

The second category departs from reliance on predefined features, claiming that handcrafted methods have demonstrated limited effectiveness in extracting generalizable and distinctive representations~\cite{nanni2017handcrafted,el2020deep}. In this line, Lu et~al.~\cite{lu2020vision} and follow-up work~\cite{lu2021quantifying} introduced an ordinal focal convolution network that concatenates coordinate vectors of all joints to estimate the UPDRS-gait score. A notable trend in this line of work involves adapting the ST-GCN motion encoder model~\cite{yan2018spatial} for skeleton-based abnormal gait analysis, (e.g,~\cite{mehta2021towards, zeng2023video, guo2021multi, sabo2022estimating, tian2022skeleton, tian2024cross}). In this line of work, Sabo et~al.~\cite{sabo2022estimating} systematically evaluated several ST-GCN-based models. %, finding that they consistently outperformed all other models and feature sets. 
More recently, Tian et~al.~\cite{tian2024cross} introduced a cross-spatiotemporal graph convolution network (CST-GCN) designed to learn cross-spacetime dependencies. Overall, these methods achieve an accuracy ranging between 55\% and 70\%. Given the nature of our 3-class classification task, where a random classifier would achieve a 33\% accuracy, it is apparent that there is room for improvement.

The improved performance obtained by using ST-GCN models as a backbone for predicting UPDRS-gait scores, along with the identified potential for improvement, raises the question of why there has been no initiative to explore alternative human motion encoder models in this clinical context. This observation motivates our exploration of state-of-the-art motion encoders, which have demonstrated excellent performance in tasks such as activity recognition and pose estimation, to predict UPDRS-gait scores in PD patients.

%Collectively, these highlighted works underscore the potential of ambient monitoring, vision-based pose estimation, and deep learning techniques in extracting meaningful gait features and predicting PD severity. This collective evidence motivates our exploration of applying state-of-the-art motion prior-based models, successful in tasks like activity recognition and pose estimation, to UPDRS-III-gait scores in PD patients.

\subsection{Human motion encoders}

The field of human motion analysis has undergone a remarkable transformation with the introduction of human motion encoder models. These models, pre-trained on extensive datasets excel in tasks like 3D human pose estimation and action recognition, capturing inherent spatiotemporal patterns within human movement.

We selected six state-of-the-art human motion encoders as the backbone for Parkinsonian score prediction based on their performance on benchmark datasets: PD ST-GCN~\cite{sabo2022estimating}, MixSTE~\cite{zhang2022mixste}, MotionAGFormer~\cite{motionagformer2024}, MotionBERT~\cite{motionbert2022}, PoseFormerV2~\cite{Zhao_2023_CVPR}, and Pose Transformers (POTR)~\cite{POTR}. These models, each with unique architectural features suitable for capturing complex motion patterns, represent the latest advances in human motion coding and have recently benchmarked general motion analysis and downstream tasks.

Despite their success, their applicability to gait analysis for PD assessment remains largely unexplored. Applying these models to gait analysis within the context of PD holds promise to surpass the limitations associated with hand-crafted feature extraction, potentially offering a more comprehensive understanding of gait impairment. However, their direct application is not without challenges, prompting the question of whether general motion encoders primarily trained on healthy populations can effectively discern nuanced differences in Parkinsonian gait severity. Our primary aim is to fill this gap by conducting a systematic assessment of state-of-the-art human motion encoders for PD severity estimation, measured by the UPDRS-gait scale. To our knowledge, no prior work has performed a comparative and transitional analysis of such models for clinical tasks.

%\cite{endo2022gaitforemer} not sure whether or where to add it?

%% file: 3_Method.tex
\vspace{10pt}
\section{METHOD}
\vspace{5pt}

In this section, we present a framework that leverages the recent human motion encoder models. This approach is designed to distill valuable insights from complex gait patterns, transforming raw skeletal data into clinically actionable UPDRS-gait scores. Our methodology encapsulates a dual approach: firstly, by exploiting the latest advancements in motion encoding to directly infer clinical scores from gait motion, and secondly, by utilizing traditional gait features as a comparative benchmark. Figure~\ref{fig:diagram} illustrates an overview of the framework. The subsequent sections delineate the problem definition and elaborate on the backbone models and gait features.

\subsection{Problem definition}

Given a raw gait sequence $g^{1:F}$, where \(g\) represents the 3D location of each joint and $F$ denotes the total number of frames in the sequence, we first apply a series of preprocessing steps $\mathcal{P}$:

% \begin{equation}
% p^{1:F'} = \mathcal{P}(g^{1:F})
% \end{equation}

\begin{equation}
\{p^{1:F'}_i\}_{i=1}^N = \mathcal{P}(g^{1:F})
\end{equation}

\noindent Function \( \mathcal{P} \) includes preprocessing steps such as skeleton transformation, downsampling, and sequence clipping to transform \( g^{1:F} \) into a set of \(N\) standardized pose sequence clip representations \(\{p^{1:F'}_i\}\), where each \(p^{1:F'}_i\) contains a different segment of the original sequence but maintains the same fixed temporal dimension \(F'\). The motion encoder \(\mathcal{E}\) then processes each pose sequence \(p^{1:F'}_i\) to extract relevant features $e_i$.

%into a standardized pose sequence representation \( p^{1:F'} \), with \( F' \) as the new temporal dimension after preprocessing.
%The motion encoding phase is then executed by a motion encoder $\mathcal{E}$ processes the preprocessed pose sequence $p^{1:F'}$ to extract relevant features:

\begin{equation}
e_i = \mathcal{E}(p^{1:F'}_i)
\end{equation}

% \begin{equation}
% \{e^{1:F'}_i\}_{i=1}^N = \{\mathcal{E}(p^{1:F'}_i)\}_{i=1}^N
% \end{equation}

 %where \( e^{1:F'} \) represents the encoded feature set that captures the gait dynamics characteristics.
Finally, these features are used by a classifier head $\mathcal{C}$ to predict the UPDRS-gait score for each sequence:
\begin{equation}
\text{Score}_i = \mathcal{C}(e_i)
\end{equation}

The final prediction for the gait sequence \( g^{1:F} \) is determined by taking a majority vote across all predictions \( \text{Score}_i \) for the sequence.

% \begin{equation}
% \text{Final UPDRS Score} = \text{MajorityVote}(\text{Predictions}^{1:F'})
% \end{equation}
\begin{equation}
\text{Final Score} = \text{mode}\left( \{\text{Score}_i\}_{i=1}^N \right)
\end{equation}

\subsection{Backbone Models}
We selected six recent human motion encoder models for evaluation: ST-GCN~\cite{yan2018spatial}, MixSTE~\cite{zhang2022mixste}, MotionAGFormer~\cite{motionagformer2024}, MotionBERT~\cite{motionbert2022}, PoseFormerV2~\cite{Zhao_2023_CVPR}, and Pose Transformers (POTR)~\cite{POTR}.
These models process human body skeleton data, aligning with the dataset available for Parkinson's disease~\cite{pdDataset} which consists of skeletal representations. %These models represent the latest advancements in human motion encoding, each with unique architectural features suitable for capturing complex motion patterns.
Our selection was informed by the models' performance on benchmark datasets such as Human3.6M~\cite{ionescu2013human3}, where they achieved top-tier results in different motion-related tasks. These models have different architectures specifically suited for processing motion data, and we provide a summarized overview of them.

\textit{MixSTE}~\cite{zhang2022mixste} uses a spatial transformer module to extract intra-frame relationships among body joints and a temporal transformer module for capturing inter-frame dependencies. Then it proposes a block that has spatial transformer followed by a temporal transformer. By repeating this block multiple times, it captures spatial-temporal correlation of the input sequence.

% \textit{MixSTE model} sequentially applies spatial and temporal transformers in a repeated block structure to capture spatial-temporal relationships.
% \begin{equation}
%     \mathcal{E}(p^{1:F'}) = \underbrace{\mathcal{E}_{\text{temp}} \circ \mathcal{E}_{\text{spat}} \circ \ldots \circ \mathcal{E}_{\text{temp}} \circ \mathcal{E}_{\text{spat}}}_{K\ \text{times}}(p^{1:F'})
% \label{eq:mixste}
% \end{equation}
% \noindent where $\mathcal{E}_{\text{spat}}$ represents the spatial transformer block that processes intra-frame joint relationships, and $\mathcal{E}_{\text{temp}}$ denotes the temporal transformer block that captures inter-frame dependencies. This composite function is applied $K$ times to the pose sequence \( p^{1:F'} \), effectively integrating spatial and temporal information throughout the sequence.

\textit{MotionBERT}~\cite{motionbert2022} proposes a two-branch block. One branch comprises a spatial transformer followed by temporal transformers, while the other branch consists of a temporal transformer followed by a spatial transformer. These two branches are merged by a weighted average of their output features. The weights are determined by multiplying the output feature of each branch with a matrix that is updated during training. This design enables MotionBERT to effectively capture and refine motion features across $K$ layers, incorporating both spatial and temporal dimensions through its dual-stream approach.

% \begin{align}
% \mathcal{E}(p^{1:F'}) = \sum_{k=1}^{K} &\left( \alpha_{k}^{\text{ST}} \cdot \mathcal{E}_{k}^{\text{temp}}\left(\mathcal{E}_{k}^{\text{spat}}(F_{k-1})\right) \right. \nonumber \\
% &\left. + \alpha_{k}^{\text{TS}} \cdot \mathcal{E}_{k}^{\text{spat}}\left(\mathcal{E}_{k}^{\text{temp}}(F_{k-1})\right) \right)
% \label{eq:motionbert}
% \end{align}

% \noindent $\mathcal{E}_{k}^{\text{spat}}$ and $\mathcal{E}_{k}^{\text{temp}}$ denote the spatial and temporal transformers at the $k$-th level of the network, respectively. $F_{k-1}$ represents the feature embeddings from the previous layer, and $\alpha_{k}^{\text{ST}}$ and $\alpha_{i}^{\text{TS}}$ are the adaptive fusion weights for the spatial-temporal and temporal-spatial streams, respectively. This formula captures the iterative process of feature refinement across $K$ layers, combining spatial and temporal information in a dual-stream approach.

\textit{MotionAGFormer}~\cite{motionagformer2024} substitutes MotionBERT's temporal-spatial branch with a novel module called GCNFormer, which replaces the attention module of transformers with Graph Convolutional Networks. The spatial GCNFormer's adjacency matrix relies on physically connected edges among human joints, while the temporal GCNFormer determines its adjacency matrix by selecting the k-most similar nodes across different time frames.

% \begin{align}
% \mathcal{E}(p^{1:F'}) = \sum_{k=1}^{K} &\left( \mathcal{E}_{k}^{\text{temp}}\left(\mathcal{S}_{k}^{\text{spat}}(F_{k-1}^{\text{Transformer}})\right) \right. \nonumber \\
% &\left. + \mathcal{G}_{k}^{\text{temp}}\left(\mathcal{S}_{k}^{\text{spat}}(F_{k-1}^{\text{GCNFormer}})\right) \right)
% \label{eq:motionagformer}
% \end{align}

% Here, $\mathcal{S}_{k}^{\text{spat}}$ represents the Spatial MetaFormer processing individual body joints as distinct tokens. $\mathcal{T}_{k}^{\text{temp}}$ and $\mathcal{G}_{k}^{\text{temp}}$ signify the Temporal MetaFormer in the Transformer and GCNFormer streams, respectively. $F_{k-1}^{\text{Transformer}}$ and $F_{k-1}^{\text{GCNFormer}}$ denote the feature embeddings from the previous layer for each respective stream. This dual approach enables the model to capture both intra-frame and inter-frame relationships effectively

\textit{POTR}~\cite{POTR} employs a non-autoregressive approach for human motion prediction, decoding all elements of a sequence in parallel. This design reduces computational demands and limits error propagation over extended sequences. The POTR architecture consists of three main components: a pose encoding neural network that computes pose embeddings for each 3D pose vector in the input sequence, a non-autoregressive Transformer for learning temporal dependencies, and a pose decoding network that generates a sequence of 3D pose vectors. The Transformer model in POTR includes both self- and encoder-decoder attention modules, allowing for the identification of relevant elements in the input sequence for prediction. Notably, for our study, we use the Transformer encoder component of the POTR model as the motion encoder.

\textit{PoseFormerV2}~\cite{Zhao_2023_CVPR} optimizes computational efficiency by using a subset of central frames as input for the spatial transformer. To capture extensive human dynamics within the input sequence, all frames undergo transformation into Discrete Cosine Transform (DCT) coefficients. A low-pass filter is then applied to retain only a specific subset of coefficients for each joint trajectory, ensuring effective representation of long-range human movements. Subsequently, the combined output tokens from the spatial transformer and low-pass filter enter a temporal transformer. Here, an attention module processes the tokens, sending those related to the frequency domain to a Multilayer Perceptron (MLP), while those associated with the time domain undergo conversion into the frequency domain using DCT. Following MLP processing, the tokens are reverted to the time domain using Inverse DCT. Note that unlike other models this flow is not repeated, and the output of temporal transformer is used as the ultimate high-level feature vector.

% \begin{align}
% \mathcal{E}(p^{1:F'}) &= \text{TF-Fusion}\left( \mathcal{E}_{\text{spat}}(p^{1:F'}_{\text{sampled}}), \mathcal{E}_{\text{freq}}(p^{1:F'}) \right) \\
% %\mathcal{E}_{\text{spat}}(p^{1:F'}_{\text{sampled}}) &= \text{SpatialTransformer}(p^{1:F'}_{\text{sampled}}) \\
% %\mathcal{E}_{\text{freq}}(p^{1:F'}) &= \text{DCT-LowPass}(p^{1:F'})
% \end{align}

% \noindent \(\mathcal{E}_{\text{spat}}\) denotes the spatial transformer that processes a central subset of frames \(p^{1:F'}_{\text{sampled}}\). \(\mathcal{E}_{\text{freq}}\) represents the encoding of the full pose sequence \(p^{1:F'}\) into frequency domain using Discrete Cosine Transform (DCT) followed by a low-pass filtering operation to retain only essential frequency components. The final encoding \(\mathcal{E}\) is achieved by the Time-Frequency Feature Fusion (TF-Fusion) module that effectively merges these spatial and frequency domain features for subsequent classification tasks

\textit{ST-GCN}~\cite{yan2018spatial} models operate on trajectories of joint positions, using filters that consider neighboring joints in the spatial dimension rather than aggregating information across all joints at each timestep. This approach leverages domain-specific knowledge about joint connectivity, enhancing the models' ability to interpret skeletal data~\cite{sato2019quantifying}.
% We used the ST-GCN model in~\cite{sabo2022estimating}, which 
We employed the best performing ST-GCN model from~\cite{sabo2022estimating}, where three ST-GCN backbone architectures were evaluated. We will consistently denote this model as ``PD~ST-GCN'' throughout the paper.

\subsection{Feature-Based Model}

This approach implements a feature-based classification method to estimate UPDRS-gait scores, using gait features extracted from the input sequence. To detect gait cycles the recently proposed Transforming Gait model~\cite{TransformingGait} is used, which takes 3D joint location sequences % estimated by pre-trained algorithms~\cite{contributors2020openmmlab, gong2021poseaug} 
and detects critical gait events such as foot contact and toe-off. These events are necessary for estimating gait features such as cadence, step time, length and width, walking speed, and margin of stability which are fundamental to the assessment of PD severity~\cite{sabo2020assessment}.
%by the Transforming Gait model~\cite{TransformingGait}.
%\subsubsection{Transforming Gait Model}
%The Transforming Gait model is trained on over 7,000 monocular video sequences from a clinical gait analysis lab. This model takes 3D joint location sequences estimated by pre-trained algorithms and outputs kinematic trajectories and timing information. 
% These outputs allow for the measurement of several gait parameters on individual gait cycles, demonstrating a high correlation with parameters from formal gait analysis.
%The Transforming Gait model detects critical gait events such as foot contact and toe-off with high precision. These events are pivotal for estimating gait features such as cadence, step time, length and width, walking speed, and Margin Of Stability (MOS) which are fundamental to the assessment of PD severity.
%The Transforming Gait model employs a transformer to map sequences of 3D joint locations and the height of the subject to a set of interpretable features for each frame. The algorithm is made invariant to camera angles by rotating the 3D joints to a standard orientation before processing. 
%The kinematic outputs of this model include hip and knee joint angles, the forward position of the foot and the velocity of the pelvis and feet.
% which are used along with the velocity of the pelvis and feet to extract foot contact timings and compute gait parameters.
The timing of heel-strike and toe-off events is encoded using quadrature encoding of the gait phase, which allows for the reconstruction of the phase from model outputs. A Kalman smoother is applied to the quadrature encoded phase signals to detect gait events robustly and reduce noise. 

% This two-step approach is divided into: Gait Features Estimation and Gait feature UPDRS-gait Classifier.
% The encoder of the transformer comprises six layers with six attention heads per layer, trained using an Adam optimizer over 150 epochs. The inclusion of dropout layers and layer normalization ensures the model's generalization across different sequences. For more details please refer to~\cite{TransformingGait}.

\subsubsection{Gait Features Estimation} We select a set of gait features for UPDRS-gait scores estimation based on extensive experiments. The most effective combination of features, yielding the best results,  includes cadence, step time, walking speed, step length, step width, and margin of stability.

% The Transforming Gait model's integration into our system design allows for the extraction of gait features that are both precise and clinically meaningful. 
% This model facilitates cycle-by-cycle gait parameter estimation, offering a granular and accurate representation of gait suitable for assessing PD.
Step width reflects gait stability and balance control, which are often impaired in individuals with PD due to rigidity and postural instability. A compromised step width can be associated with an increased risk of tripping and falling, a prevalent issue in PD.
Step width is calculated as the lateral absolute distance between the feet during a step. Step length is calculated similarly but uses the anteroposterior (AP) direction. The estimation of step time is based on the difference in heel-strike times divided by the frame rate (fps). Walking speed is defined as the total distance covered by the sacral marker over the duration of walk, divided by the total time taken. Cadence refers to the number of steps taken in a minute. Margin of stability, defined as the minimum distance between the extrapolated center of mass (XCOM) and the base of support~\cite{mehdizadeh2022toronto}, is a direct indicator of postural stability. In PD patients, margin of stability is generally reduced, indicating a lower ability to maintain balance, particularly during gait transitions. 
For each walk, we compute the mean and standard deviation of features (a measure of variability) across all observed steps, except for MOS where we use the minimum instead of the mean.

\subsubsection{Gait feature UPDRS-gait Classifier}

The extracted gait features serve as inputs to the UPDRS-gait classifier. We train a Random Forest classifier that takes as input cadence, walking speed, step length, step width, step time, and margin of stability along with the variability measures and outputs the UPDRS-gait score.

    \begin{table*}[!t]\normalsize
    \caption{Weighted average performance of feature-based model and different motion encoder-based models in estimation UPDRS-gait scores.}
      \centering
        \begin{tabular}{l|cccc}
          \hline
          Method & Accuracy & Precision & Recall & F1-Score
     \\
          \hline
          Feature-based & \textbf{0.66} & \textbf{0.68} & \textbf{0.66} & \textbf{0.66} \\
          \hline
          MixSTE (CVPR2022) & 0.40 & 0.41 & 0.40 & 0.41 \\
          MotionAGFormer (WACV2024) & 0.42 & 0.42 & 0.42 & 0.42 \\
          MotionBERT-LITE (ICCV2023) & 0.42 & 0.45 & 0.42 & 0.43 \\
          POTR (ICCV2021) & 0.45 & 0.49 & 0.45 & 0.46 \\
          MotionBERT (ICCV2023) & 0.48 & 0.47 & 0.48 & 0.47 \\
          PD STGCN (JBHI2022) & 0.49 & 0.50 & 0.49 & 0.48 \\
          PoseFormerV2 (CVPR2023) & {0.60} & {0.61} & {0.60} & {0.59} \\
          PoseFormerV2-Finetuned & \underline{0.64} &  \underline{0.65} &  \underline{0.64} &  \underline{0.62} \\
            
          \hline
        \end{tabular}%
    \label{tab:encoder-comparison}
    \end{table*}

\subsection{Data}

We performed our experiments using a recently released clinical dataset~\cite{pdDataset}, the first public full-body kinematics and kinetics gait analysis dataset for Parkinson's Disease. The dataset includes 885 walking sequences from 26 idiopathic individuals with PD captured using motion-capture sensors. Our study, the first to use this dataset for PD severity estimation, employs UPDRS-gait annotations, scored by two experienced physiotherapists. The dataset includes 472 ON and 413 OFF medication walks from each participant, which we use to evaluate the sensitivity of predictive models to clinically relevant changes.
Data from 3 participants are excluded from our study due to motion capture issues. It is important to note that participants are assigned a single UPDRS-gait score for all their walks in an ON (or OFF) medication state. Consequently, our models' walk-specific predictions are compared against a single label for each medication state, which potentially oversimplifies the variability in gait patterns within each state.

\subsection{Data Preparation}
A LOSOCV strategy is implemented to develop and evaluate the predictive models. The training set comprises data from 16 participants, while the validation set includes data from 6 other participants, with an even distribution across UPDRS scores 0, 1, and 2, for model tuning and validation. The test set consists of data from the remaining 1 participant, used to assess the models' performance on unseen data. This setup results in a 23-fold validation process, where each participant's data is used once as the test set. 

We perform several preprocessing steps (shown in Figure~\ref{fig:diagram}-right). First, the motion capture data of the PD dataset, consisting of 44 joints, is transformed into the standardized 17-joint Human3.6M~\cite{ionescu2013human3} motion format (Figure~\ref{fig:diagram}-Skeleton Transformation). To achieve this, we perform interpolation for certain joints, estimating some of the joints from the others.
The dataset features high-resolution walking sequences at 100~FPS. To adapt this data for state-of-the-art motion encoders, which are typically trained on 30~FPS data, the sequences are downsampled to match this frequency. Also, considering the extensive duration of the recorded sequences and the input constraints of the motion encoders, each video is segmented into clips consisting of a fixed length of 81 frames. 
% MotionBERT expands these clips to 243 frames, while others retain the original 81 frames. 
After processing these fixed-length clips through the motion encoders and PD classifiers, a majority voting approach was employed to determine the final estimation for the entire sequence. 

%% file: 4_Experiments.tex
\vspace{10pt}
\section{EXPERIMENTS}
\vspace{5pt}
    
    In this section, we detail our experiments and evaluation results on the PD dataset and provide an in-depth discussion of the results. %of each model's performance.%, including a comparison with feature-based models.

\subsection{Experimental Settings}
    Each motion encoder model is evaluated using a linear classifier to assess the baseline performance of each method without any additional enhancements or complexity. Hyperparameter tuning (for the backbone and classifier) is conducted separately for each model to ensure optimal and fair performance. 
    The hyperparameter configurations for the best-performing models (learning rate, type of optimizer and drop out for classifiers, and the hyper-parameters specific to each motion encoder model) are detailed in the code\footnote{\href{https://github.com/TaatiTeam/MotionEncoders_parkinsonism_benchmark.git}{https://MotionEncoders-parkinsonism-benchmark/}}. To address data imbalance, we use a weighted cross-entropy loss. Given the limited size of the dataset, four data augmentations are employed to prevent overfitting: random rotations, random noise addition, motion mirroring, and random axis masking. For models, except for PD~STGCN and POTR, 2D joint locations derived from 3D data are used, aligning with their original training on 2D to 3D lifting tasks. 
    For MotionAGFormer, MixSTE, and PoseFormerV2, we chose the small version (accepting 81 frames as input) because the larger version requires 243 frames, which would exclude many of our gait sequences due to their shorter length. However, for MotionBERT, we evaluate both the small and large versions, as it can process shorter videos by extrapolating them to 243 frames. For POTR and PD STGCN, we use the original version designed for 80 frames. %Each of these augmentations enhances the robustness of the predictive model without compromising the integrity of the Parkinson's Disease dataset.

    Additionally, a hyperparameter search is conducted for the feature-based model with a Random Forest classifier. Other classification models, including SVM and XGBoost, as well as various gait feature combinations were evaluated. However, these models yielded slightly worse results in preliminary experiments and, for brevity, are not presented.

    The models' performance in estimating UPDRS-gait severity is evaluated using accuracy, weighted average precision, recall, and F1-Scores to account for the dataset class imbalance effect. Furthermore, we perform a Wilcoxon ranked test to analyze the significance of UPDRS-gait predictions between patients' on and off medication states.

\subsection{Experimental Results and Discussion}

Table~\ref{tab:encoder-comparison} presents the results of the feature-based model and motion encoder-based models in estimating UPDRS-gait scores. While the feature-based model achieves an F1-score of 0.66, it is hindered by the need for extensive manual corrections. We found that the Transforming Gait model sometimes generated inaccurate heel strikes, particularly in severe PD cases. We had to correct these inaccuracies manually, yet there remains uncertainty about the accuracy of estimated gait features. This not only underscores the limitations of such models, but also highlights the intensive data cleaning and feature engineering required for each task. In contrast, automated motion encoder models, like PoseFormerV2, with their comparable performance, though not yet surpassing the feature-based approach in all metrics, offer promising alternatives for capturing gait dynamics more efficiently, minimizing the need for manual corrections. %Among the motion encoder-based models, PoseFormerV2~\cite{Zhao_2023_CVPR} and PoseFormerV2-Finetuned show superior performance, with the fine-tuned version achieving the highest scores. This suggests that fine-tuning on task-specific data is critical for model performance. 
POTR~\cite{POTR}, PD~STGCN~\cite{sabo2022estimating}, and MotionBERT~\cite{motionbert2022} also show promising results but underperform PoseFormerV2 and the feature-based model. 
% with their higher precision indicating a potential for capturing the nuanced patterns of PD gait more effectively than other encoder models.
Notably, models like MixSTE~\cite{zhang2022mixste} and MotionAGFormer~\cite{motionagformer2024}, despite being state-of-the-art for 3D pose lifting, did not perform as well as the feature-based model. This could imply that without task-specific fine-tuning, even advanced models may struggle to adapt to the specialized requirements of clinical gait analysis.

We emphasize that the results reported in Table~\ref{tab:encoder-comparison} are based on LOSOCV. In some published research~\cite{sabo2020assessment}, standard cross-validation (CV) is employed instead of LOSOCV. In Table~\ref{tab:standard-splitting}, we present PoseFormerV2-Finetuned results using standard CV with a 70/15/15 split for training, validation, and testing.
% Additionally, to compare our results with methods that used walk splits~\cite{sabo2020assessment} instead of a held-out participant division, we experimented with a 70/15/15 split for training, validation, and testing. The results, presented in Table~\ref{tab:standard-splitting}, show performance with this conventional split. 
A notable performance decrease in LOSOCV compared to standard CV (F1-score of 0.62 vs 0.75) highlights the risk of overestimating model performance with non-participant-specific data divisions. This emphasizes the presence of significant inter-person variations in gait sequences, and models show improved performance when they see a few walks from the individual on whom they are being tested. Our use of the LOSOCV approach provides a more realistic assessment of model effectiveness in real-world applications.
Given the absence of prior work on parkinsonian score prediction using this dataset, and the lack of publicly available datasets from other studies, direct comparison is not feasible. However, contextualizing the F1-score of 0.62 from this paper alongside similar studies using skeleton data, such as \cite{sabo2022estimating} with an F1-score of 0.52, demonstrates good performance.

\begin{table}[!t]\small
    \caption{Weighted average performance of PoseFormerV2-Finetuned using standard cross-validation (instead of LOSOCV).}
      \centering
        \begin{tabular}{l|ccc}
          \hline
            & Precision & Recall & F1-Score
     \\ \hline
          UPDRS-gait Score: 0 & 0.83 & 0.78 & 0.80 \\
          UPDRS-gait Score: 1 & 0.66 & 0.72 & 0.69 \\
         UPDRS-gait Score: 2 & 0.75 & 0.73 & 0.74 \\
          % Accuracy &  &  & 0.75 \\
          Weighted Avg & 0.75 & 0.75 & 0.75 \\
            
          \hline
        \end{tabular}%
    \label{tab:standard-splitting}
    \end{table}

Furthermore, to determine whether there were significant differences in model UPDRS-gait predictions between patients' ON and OFF medication, we applied the Wilcoxon signed-rank test to explore the models' sensitivity to medication-induced changes. The expectation is that the scores would be higher in OFF medication states. Table~\ref{tab:onoff} presents the Wilcoxon signed-rank test statistics and corresponding p-values for the comparison between ground-truth and various models' estimations of UPDRS-gait scores across ON and OFF medication states. %The asterisks indicate the levels of significance, with single and double asterisks denoting p-values below 0.05 and 0.01, respectively.

The results demonstrate that several models, including PoseFormerV2, PoseFormerV2-Finetuned, and MotionBERT, have statistically significant differences in performance between ON and OFF medication states, indicating sensitivity to medication-induced changes in gait. As expected, ground truth (i.e. experienced physiotherapists) scores were significantly different for the ON and OFF medication states (p-value=0.0198). Furthermore, statistically significant differences ($p < 0.05$) were observed between ON and OFF model predictions of MotionAGFormer, MotionBERT, MixSTE, PoseFormerV2, and PoseFormerV2-Finetuned. Notably, the p-value for PoseFormerV2-Finetuned was smaller than 0.01.
%In contrast, models like POTR and MotionBERT-LITE did not show significant results, suggesting they might be less sensitive to the distinctions between medication states or require further optimization for this specific task. 
% The feature-based model approached significance ($p = 0.0641$), suggesting that traditional features still capture some medication-related gait changes, although not as strongly as the best-performing motion encoders. 
The statistical analysis underscores the potential of motion encoder models, particularly those fine-tuned like PoseFormerV2-Finetuned~($p < 0.01$), in being sensitive enough to detect clinically relevant changes in gait patterns.

\begin{table}\small
    \caption{Wilcoxon signed-rank statistics and p-values for ground-truth and different models based on ON and OFF medication states.}
      \centering
        \begin{tabular}{l|cc}
          \hline
          Method & Test Statistic & p-value
     \\
          \hline
           Ground-truth & 32.0 & .0198* \\
          \hline
          Feature-based & 9.0 & .0037**\\
          \hline
          MotionBERT-LITE (ICCV2023) & 2.5 & .7854\\
          POTR (ICCV2021) & 52.0 & .2460\\
          PD STGCN (JBHI2022) & 27.0 & .0542 \\
          MotionAGFormer (WACV2024) & 17.0 & .0463*\\
          MotionBERT (ICCV2023) & 13.0 & .0409*\\
          MixSTE (CVPR2022) & 16.0 & .0392*\\
          PoseFormerV2  (CVPR2023) & 26.0 & .0168*\\
          PoseFormerV2-Finetuned  & 14.0 & .0052**\\
            
          \hline 
        \end{tabular}%
    \label{tab:onoff} \\
    \vspace{3pt}
    \footnotesize{* significant at p $<$ 0.05, ** significant at p $<$ 0.01}
    \end{table}

% \begin{figure*}[t]
%   \centering
%   \begin{subfigure}[b]{0.32\linewidth} % Adjust the width as needed to fit in a single row
%     \includegraphics[width=\linewidth]{images/best_confusion_matrix_allfolds_Posev2.png}
%     % \caption{\footnotesize Feature-based}
%     \label{fig:confusions_Feature-based}
%   \end{subfigure}\hfill
%   \begin{subfigure}[b]{0.32\linewidth} % Adjust the width as needed to fit in a single row
%     \includegraphics[width=\linewidth]{images/best_confusion_matrix_allfolds_OFF_Posev2.png}
%     % \caption{\footnotesize Motion prior-based}
%     \label{fig:confusions_prior-based_off}
%   \end{subfigure}\hfill
%   \begin{subfigure}[b]{0.32\linewidth} % Adjust the width as needed to fit in a single row
%     \includegraphics[width=\linewidth]{images/best_confusion_matrix_allfolds_ON_Posev2.png}
%     % \caption{\footnotesize Motion prior-based}
%     \label{fig:confusions_prior-based_on}
%   \end{subfigure}
%   \caption{Confusion matrices for UPDRS-gait score predictions from PoseFormerV2-Finetuned overall (left), OFF (middle) and ON (right) medication states. }
%   \label{fig:confusions}
% \end{figure*}

\begin{figure*}[t]
  \centering
  \begin{subfigure}[b]{0.25\linewidth} 
    \includegraphics[width=\linewidth]{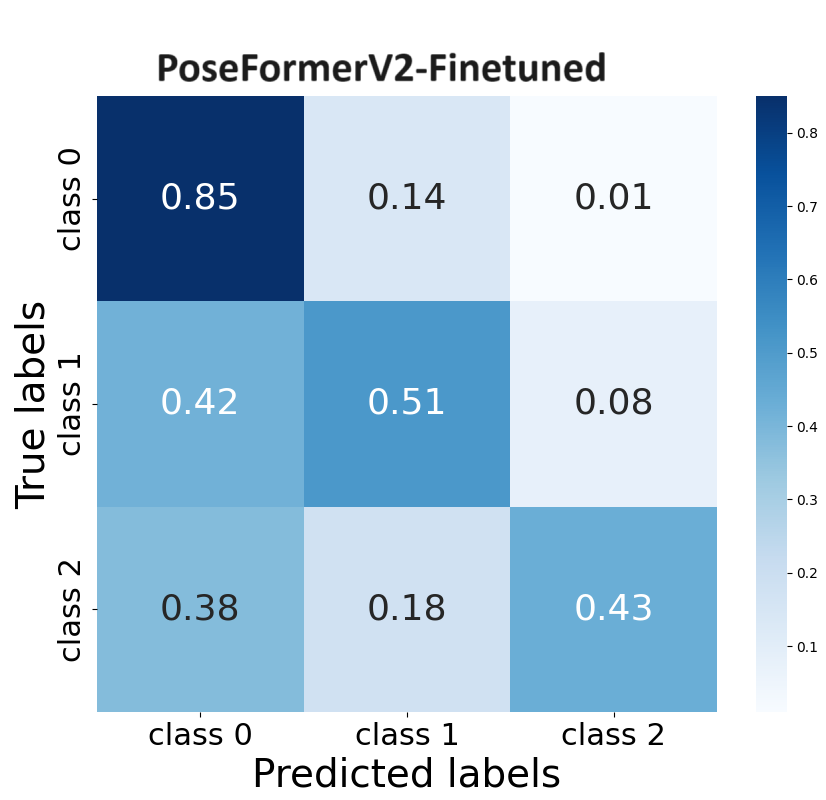}
    % \caption{\footnotesize Feature-based}
    \label{fig:confusions_Feature-based}
  \end{subfigure} \hspace{1cm}
  \begin{subfigure}[b]{0.25\linewidth} 
    \includegraphics[width=\linewidth]{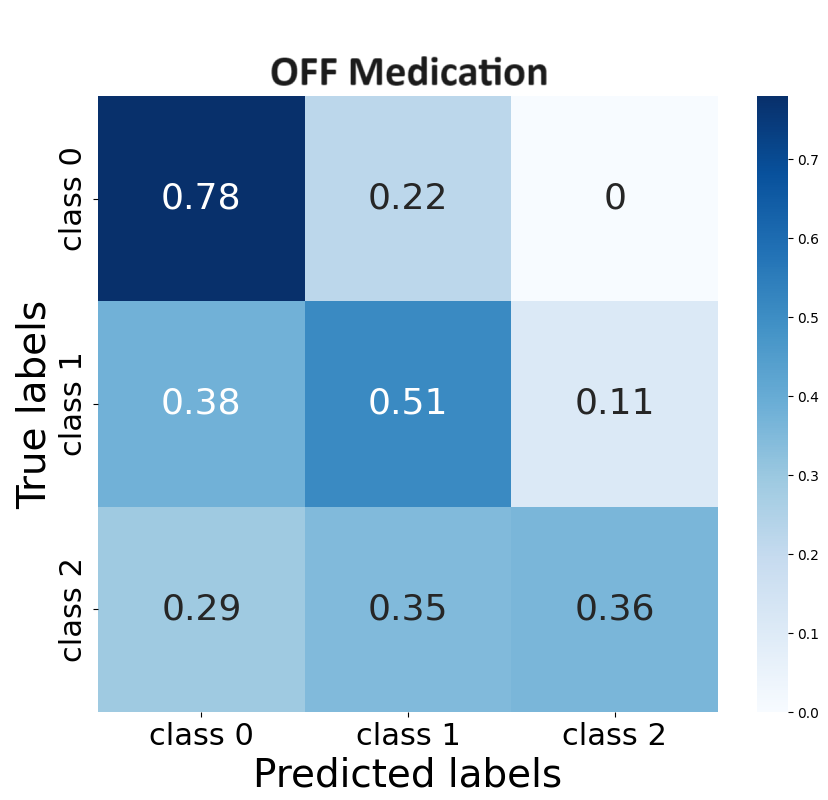}
    % \caption{\footnotesize Motion prior-based}
    \label{fig:confusions_prior-based_off}
  \end{subfigure} \hspace{1cm}
  \begin{subfigure}[b]{0.25\linewidth} 
    \includegraphics[width=\linewidth]{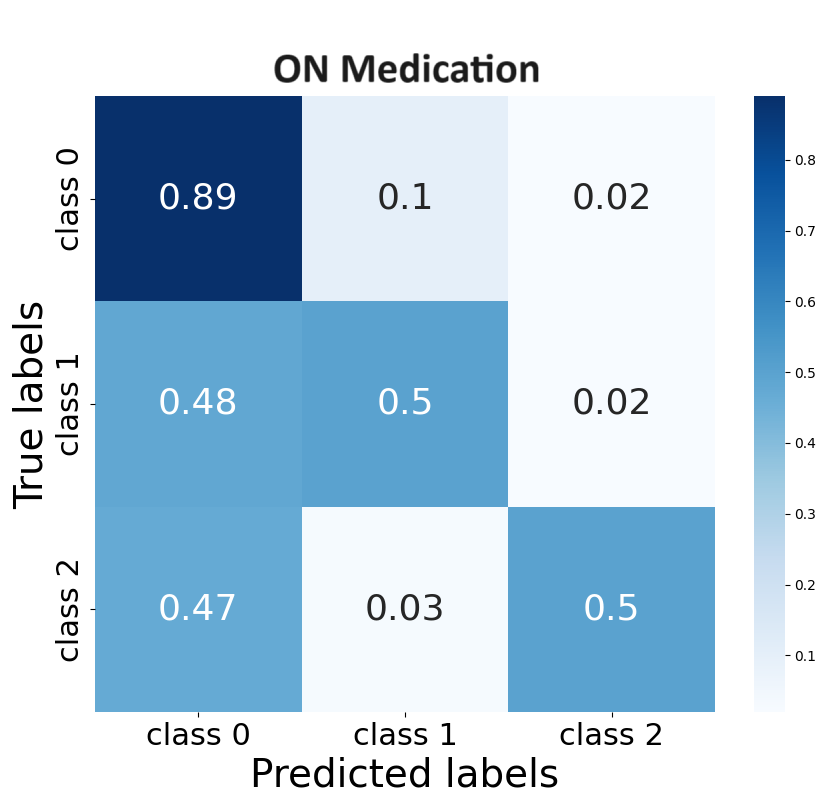}
    % \caption{\footnotesize Motion prior-based}
    \label{fig:confusions_prior-based_on}
  \end{subfigure}\\
  \begin{subfigure}[b]{0.25\linewidth} 
    \includegraphics[width=\linewidth]{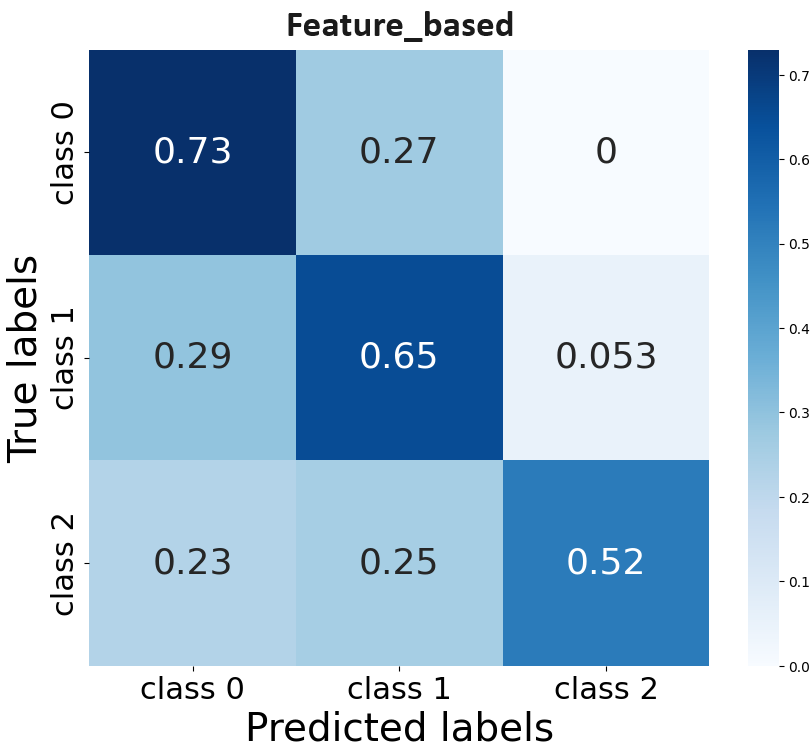}
    % \caption{\footnotesize Feature-based}
    \label{fig:confusions_Feature-based}
  \end{subfigure}\hspace{1cm}
  \begin{subfigure}[b]{0.25\linewidth} 
    \includegraphics[width=\linewidth]{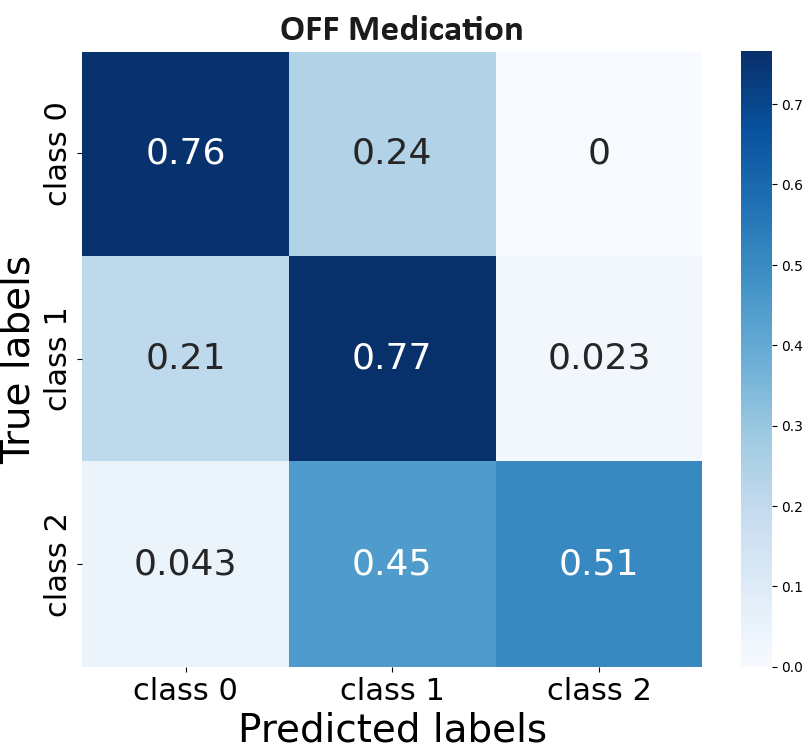}
    % \caption{\footnotesize Feature-based}
    \label{fig:confusions_Feature-based}
  \end{subfigure}\hspace{1cm}
  \begin{subfigure}[b]{0.25\linewidth} 
    \includegraphics[width=\linewidth]{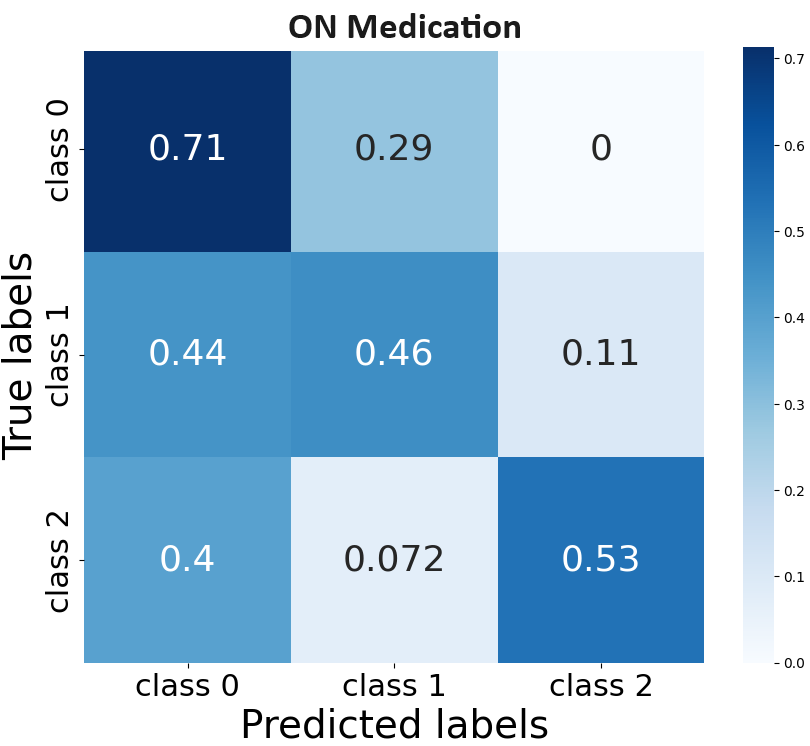}
    % \caption{\footnotesize Feature-based}
    \label{fig:confusions_Feature-based}
  \end{subfigure}
  \caption{Normalized confusion matrices for UPDRS-gait score predictions from PoseFormerV2-Finetuned (top row) and feature-based models (bottom row) across overall (left), OFF (middle), and ON (right) medication states. }
  \label{fig:confusions}
\end{figure*}

Figure~\ref{fig:confusions} illustrates the confusion matrices (normalized such that the numbers for each ground-truth class add up to 1) for UPDRS-gait score predictions from the PoseFormerV2-Finetuned and feature-based models on all data and separately for patients in the OFF and ON medication states. 
The PoseFormerV2-Finetuned shows a strong ability to correctly predict class 0 (no gait disability), but it seems to struggle with a higher rate of misclassification between classes 1 (mild gait disability) and 2 (severe gait disability). In contrast, the feature-based model exhibits a balanced performance across all classes.
%Figure~\ref{fig:confusions} illustrates the same confusion matrices but normalized such that the numbers for each ground-truth class add up to 1. 
When examining the OFF medication state, there is a notable variability in the predictions, with an increased misclassification between classes 1 and 2. This could suggest that gait patterns become less distinct or that the model is less able to differentiate between severity levels when patients are OFF medication.
Conversely, the PoseFormerV2-Finetuned ON medication state shows improved classification for class 1, yet some confusion remains between classes 0 and 2. The improvement in class 1 accuracy may reflect the model's sensitivity to medication-induced changes in gait dynamics, which could become more pronounced when patients are ON medication.

%Figure~\ref{fig:Spaghetti} compares 
% We also compared UPDRS-gait score changes for each patient ON and OFF medication in ground-truth versus model-predicted changes. We found that the model captures the trend of UPDRS score increase when patients are off medication, indicating its potential usefulness in clinical settings. 
% 
% The ground-truth demonstrates the expected variability in UPDRS score changes due to medication, a characteristic of individualized responses in Parkinson's disease treatment. Some ground-truth trajectories even suggest score reductions when OFF medication, which may reflect the natural fluctuations in disease symptoms or inconsistencies in self-reporting. The model captures these trends to some degree, evidenced by its predictions that generally show an increase in UPDRS scores when patients are OFF medication, consistent with typical symptom progression. However, the model's predictions vary in accuracy, suggesting a need for improved modeling of individual responses to medication.

% \begin{figure}
%   \centering
%   \includegraphics[width=0.9\linewidth]{images/Spaghetti-col.pdf}
%   \caption{UPDRS score dynamics for each patient ON and OFF medication: Ground-truth (top) and prediction outcomes (bottom).}
%   \label{fig:Spaghetti}
% \end{figure}